\pdfoutput=1

\documentclass[11pt]{article}

\usepackage{emnlp2021}

\PassOptionsToPackage{hidelinks}{hyperref}

\usepackage{times}
\usepackage{latexsym}
\usepackage{times}
\usepackage{soul}
\usepackage{url}
\usepackage{amssymb}
\usepackage[hidelinks]{hyperref}

\usepackage[utf8]{inputenc}

\usepackage{caption}
\usepackage{graphicx}
\usepackage{amsmath}
\usepackage{multirow}
\usepackage{amsthm}
\usepackage{amsmath}
\usepackage{booktabs}
\usepackage{algorithm}
\usepackage{algorithmic}
\usepackage{booktabs}
\usepackage{threeparttable}

\usepackage[T1]{fontenc}

\usepackage[utf8]{inputenc}

\usepackage{microtype}

\title{Gumbel-Attention for Multi-modal Machine Translation}

\author{
Pengbo Liu$^1$
\and
Hailong Cao$^1$\and
Tiejun Zhao$^1$ \\

$^1$Harbin Institute of Technology, Harbin, China\\

liupengbo.work@gmail.com, \{hailong,tjzhao\}@mtlab.hit.edu.com
}

\begin{document}
\maketitle
\begin{abstract}
Multi-modal machine translation (MMT) improves translation quality by introducing visual information. However, the existing MMT model ignores the problem that the image will bring information irrelevant to the text, causing much noise to the model and affecting the translation quality. This paper proposes a novel Gumbel-Attention for multi-modal machine translation, which selects the text-related parts of the image features. Specifically, different from the previous attention-based method, we first use a differentiable method to select the image information and automatically remove the useless parts of the image features. 
Experiments prove that our method retains the image features related to the text, and the remaining parts help the MMT model generates better translations.
\end{abstract}

\section{Introduction}

Multi-modal machine translation (MMT) is a novel research field of machine translation, which considers text information and uses visual information.
Recent research explores various methods based on the seq2seq network for MMT. 
~\citet{DBLP:conf/wmt/HuangLSOD16}transform and make the image features as one of the steps in the encoder as text in order to make it possible to attend to both the text and the image while decoding.
~\citet{DBLP:conf/emnlp/ZhouCLY18} adopts the mechanism of visual attention to jointly optimize the shared visual language embedding and model, which links visual semantics with corresponding text semantics. 
Recently, ~\citet{DBLP:conf/acl/YaoW20} models text modal and vision modal from the perspective of a graph network based on Transformer.
Previous methods integrate multi-modal information in multi-modal machine translation, and has made great progress. 
Despite their success, the current studies did not exploit how to automatically select the valuable information to the text in the image mixed with noise.
The noise in the image will not help understand the context and affect the performance of the model.

\begin{figure}[t]
	\centering\includegraphics[width=0.48\textwidth]{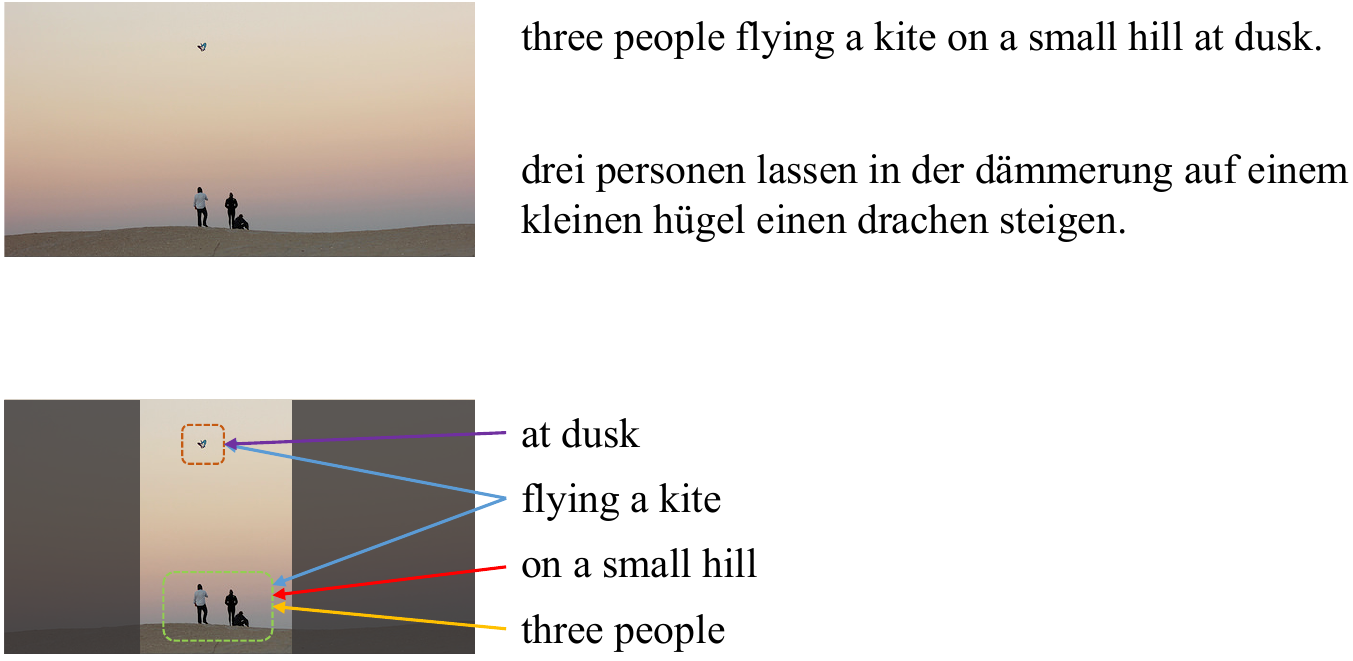}
	\footnotesize\caption{An example of useless information in the Multi30k dataset. The above one is the original picture, and the bottom is the remaining part after masking the parts irrelevant to the text in the picture.} 
	\label{fig:01}
\end{figure}

As shown in Figure 1, more than two-thirds of the original image is masked, and the remaining part can still fully express the semantics of the text. Even if we only keep the area around the entities in the picture (such as ``three people", ``kites"), we can explain all the content in the text. 
Namely, most of the content in the image is not related to the text in some scenarios, and a selecting method is essential.

In order to solve the above issues, we propose a novel attention mechanism using  Gumbel-Sigmoid~\citep{DBLP:conf/acl/GengWWQLT20} to automatically select meaningful information in the image, called Gumebl-Attention.
To the best of our knowledge, this is the first attempt to perform denoising with selecting method during training among multimodal machine translation models.
We also design a loss function to constrain the visual representation, expecting the image-aware text representation to be semantically similar to the text representation. It further ensures that parts irrelevant to the text in the image are removed.
Experiments show that our method achieves or nears the state-of-the-art performance on the three test datasets.

\section{Approach}

Our model is based on the structure of Transformer and incorporates visual modal information. 
In this section, we will elaborate on our proposed Gumbel-Attention MMT model.

\subsection{Gumbel-Attention}

In the encoder block of multi-modal machine translation, we propose Gumbel-Attention select the regions in the image that are helpful for understanding the text semantics instead of directly using the entire picture as in the previous work. 

\textbf{Selecting and Gumbel-Sigmoid.} Selecting relevant content in the picture is a question of choosing a few elements in some candidate sets. 
The usual approach is to normalize them using the softmax function first and then select the candidate elements according to the probability. This approach is also a standard method for classification tasks. 
However, the image selection is an intermediate step of the entire model in this task, which should be differentiable, and parameters can be updated through backpropagation.

 ~\citet{DBLP:conf/acl/GengWWQLT20} proposed Gumbel-Sigmoid to select a subset of elements from all elements, helping contribute to the meaning of the sentence by paying more attention to content words. The implement of Gumbel-Sigmoid is similar to Gumbel-Softmax~\cite{DBLP:conf/iclr/JangGP17} by adding Gumbel noise in the sigmoid function:
\begin{equation}
\begin{aligned}
\begin{array}{l}\operatorname { Gumbel-Sigmoid }\left({E}_{s}\right) \\ \qquad \begin{aligned}&= \operatorname{sigmoid}\left(\left({E}_{s}+{G}^{\prime}-{G}^{\prime \prime}\right) / \tau\right) \\ 
&=\frac{\exp \left(\left({E}_{s}+{G}^{\prime}\right) / \tau\right)}{\exp \left(\left({E}_{s}+{G}^{\prime}\right) / \tau\right)+\exp \left({G}^{\prime \prime} / \tau\right)}, \end{aligned}\end{array}
\end{aligned}
\end{equation}
where the input $E_s$ is a matrix, and ${G}^{\prime}$, ${G}^{\prime \prime}$ are two independent noises called Gumbel noise. We can obtain a differentiable sample by Gumbel-Sigmoid. 
$\tau \in (0, \infty)$ is a hyperparameter controlling the distribution tendency of sampling results. When $\tau$ is smaller(such as 0.1), the sampling result tends to be closer to a real one-hot vector. In contrast, the sampling result will be more similar in each dimension when $\tau$ becomes larger.
Similarly, in the spatial features of the image, a similar method can also be used to select whether each feature is retained.

\begin{figure}[t]
	\centering\includegraphics[width=0.45\textwidth]{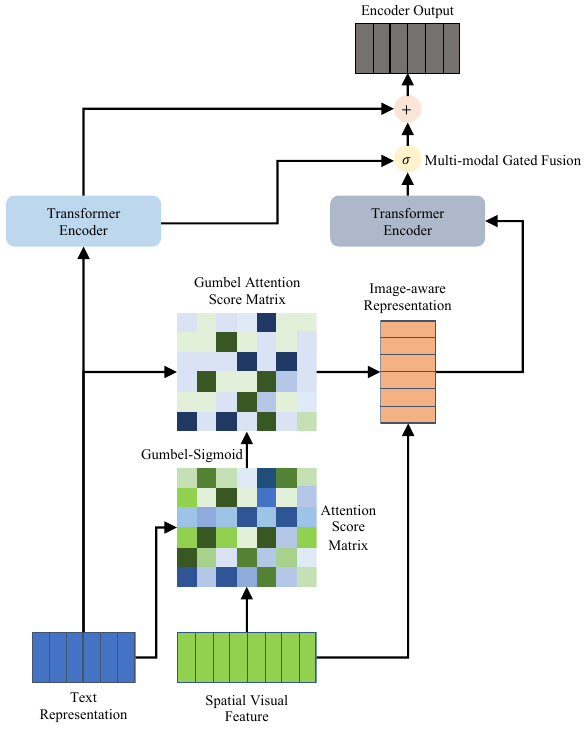}
	\footnotesize\caption{The Encoder of Our Proposed Model} 
	\label{fig:02}
\end{figure}

\textbf{Gumbel-Attention.} In order to map visual information to the semantic space of the text, the usual approach is to use the attention mechanism. In our work, we introduce an improved method for attention mechanism using Gumbel-Sigmoid, called Gumbel-Attention. 
The Gumbel-Attention mechanism can be regarded as a selection of the part of the image related to the text, which is a differentiable, discretely distributed approximate sampling. 
Similar attention mechanisms~\citep{DBLP:conf/mm/ZhengLLZ020}  have recently appeared in other fields and have achieved success.
The selecting for each element in the score matrix in the attention is as follows:

\begin{table*}[tp]

  \centering
  \begin{threeparttable}
  \setlength{\tabcolsep}{0.75mm}{
  \label{tab:performance_comparison}
    \begin{tabular}{lcccccc}
    \toprule
    \multirow{2}{*}{Model}&
    \multicolumn{2}{c}{Test2016}&\multicolumn{2}{c}{Test2017}&\multicolumn{2}{c}{MSCOCO}\cr
    \cmidrule(lr){2-3} \cmidrule(lr){4-5} \cmidrule(lr){6-7}
    &BLEU&METEOR&BLEU&METEOR&BLEU&METEOR\cr
    \midrule
    Text-only Transformer~\citep{DBLP:conf/nips/VaswaniSPUJGKP17}&37.8&55.3&29.1&48.6&25.2&44.1\cr
    Doubly-attention~\citep{DBLP:conf/acl/CalixtoLC17}&36.5&55.0&-&-&-&-\cr
    Fusion-conv~\citep{DBLP:conf/wmt/CaglayanABGBBMH17}&37.0&57.0&29.8&51.2&25.1&46.0\cr
    Trg-mul~\citep{DBLP:conf/wmt/CaglayanABGBBMH17}&37.8&55.7&30.7&{\bf 52.2}&26.4&{\bf 47.4}\cr
    Latent Variable MMT~\citep{DBLP:conf/acl/CalixtoRA19}&37.7&56.0&30.1&49.9&25.5&44.8\cr
    Deliberation networks~\citep{DBLP:conf/acl/IveMS19}&38.0&55.6&-&-&-&-\cr
    Multi-modal Transformer~\citep{DBLP:conf/acl/YaoW20}&38.7&55.7&-&-&-&-\cr
    \midrule
    Gumbel-Attention MMT&{\bf 39.2}&{\bf 57.8}&{\bf 31.4}&{ 51.2}&{\bf 26.9}&{ 46.0}\cr
    \bottomrule
    \end{tabular}}
    \end{threeparttable}
    \caption{Experimental results on the Multi30k test set. Best results are highlighted in bold.}
\end{table*}

\begin{equation}
\alpha_{i j}=\operatorname{G-S}\left(\frac{\left(x_{i}^{{text}} W^{Q}\right)\left(x_{j}^{{image}} W^{K}\right)^{T}}{\sqrt{d_{{model}}}}\right),
\end{equation}
where G-S is is the abbreviation of Gumbel-Attention, $x_{i}^{{text}}$ is the $i$ th text features, $x_{j}^{{image}}$ is the $j$ th image spatial features, $d_{{model}}$ is dimension of model, and $W^{Q} \in \mathbb{R}^{ d_{{text}} \times d_{{model}}}$, $W^{K} \in \mathbb{R}^{ d_{{image}} \times d_{{model}}}$ is trainable parameter matrices. The score-matrix consists of the sampling results of each word and all regional features of the image. Then, we will get image-aware text representation using score matrix:

\begin{equation}
\begin{aligned}
v_{i}=\sum_{j=1}^{n} \tilde{\alpha}_{i j}\left(x_{j}^{{text}} W^{V}\right),
\end{aligned}
\end{equation}
where $v_{i}$ is $i_{{th}}$ image-aware text representation which calculate the weighted sum and only use the image features related to the current word. 
To enhance the selecting accuracy of Gumbel-Attention, we also use multiple heads to improve ability of Gumbel-Attention to filter image features, just like the attention in vanilla transformer.

\subsection{Gated Fusion}

To merge the output of the two modalities,  we introduce a gating mechanism to control the fusion of $ h ^ {{image}} $ and $ h ^ {{text}} $ refering to recent work~\citep{DBLP:conf/iclr/0001C0USLZ20}:

\begin{equation}
\begin{aligned}
{\lambda}&=\operatorname{sigmoid}\left({W}h ^ {{image}} + {U}h^ {{text}}\right), \\
H&=h^{{text}} + {\lambda}h^{{image}},
\end{aligned}
\end{equation}
where $W$ and $U$ are trainable parameters. The final output $H$ is directly fed to the vanilla transformer decoder.
The encoder of the model is shown in Figure 2, and the decoder is the same as the vanilla transformer.

\subsection{Multi-modal Similarity Loss}
We propose a loss function to make the text and image expression more similar, which is also a supplement to image denoising:

\begin{equation}
\begin{aligned}
 \mathcal{L}_{sim} = \operatorname{max}(0, 1-\operatorname{cosine}(h ^ {{image}}, h ^ {{text}})-m),
\end{aligned}
\end{equation}
where ${m}$ is a hyperparameter that controls the degree of similarity margin\footnote{${m}$ should be a number from -1 to 1, 0-0.5 is suggested. In our experiment, we choose 0.3 as the value of margin.}, and we use cosine similarity to measure the similarity of two vectors. The loss function of the entire model is as follows:
\begin{equation}
\begin{aligned}
 \mathcal{L}_{total} =  \mathcal{L}_{origin} + {\alpha}\mathcal{L}_{sim},
\end{aligned}
\end{equation}\
where ${\alpha}$ is a hyperparameter that controls the proportion of $\mathcal{L}_{sim}$.
$\mathcal{L}_{total}$ is obtained by the weighted sum of the multi-modal similarity loss and the original loss.

\section{Experiment}

\subsection{Datasets}
Following previous work, we conduct experiments on the widely used  Multil30k~\citep{DBLP:conf/acl/ElliottFSS16} dataset to investigate the effectiveness of our proposed model. 
The dataset contains 29,000 instances for training, 1,014 instances for validation, and 1,000 instances for testing(Test2016). We also evaluate our model on the WMT17 test set(Test2017) and the MSCOCO test, which contain 1,000 and 461 instances, respectively.

\subsection{Setup}
Our model is implemented based on OpenNMT-py toolbox~\citep{DBLP:conf/acl/KleinKDSR17}. Due to the small size of the training corpus, our encoder and decoder only use four layers, the number of attention heads is 4, and the input and output dimensions are both 128. We adopt Adam with $\beta_1=0.9$, $\beta_2=0.98$ to optimize our model.
Spatial features are extracted from the VGG19 network as the visual representation. The feature dimension is $7 \times 7 \times 512$ , which represents the spatial information in the image. The text representation is the randomly initialized word embedding. 
Finally, we evaluate the translation quality using BLEU~\citep{DBLP:conf/acl/PapineniRWZ02} and METEOR~\citep{DBLP:conf/wmt/DenkowskiL11}.

\subsection{Results}

Table 1 shows the results of all methods on three test sets. Our Gumbel-Attention MMT model is better than most existing models, except for Trg-mul~\citep{DBLP:conf/wmt/CaglayanABGBBMH17} in meteor. One possible reason is that the result of Trg-mul comes from the system on the latest technology WMT2017 test set, which has been selected based on METEOR. An obvious finding is that our model surpasses the text-only Transformer by above 1.5 bleu points. In fact, the Gumbel-Attention-based method has only a minor modification on the vanilla Transformer, which proves the effectiveness of our model. Moreover, we draw two important conclusions:

First, compared with the Multi-modal Transformer, the main improvement of our method is the Gumbel-Sigmoid operation in the score matrix and the multi-modal similarity loss. The results of Test2016 show that this method of information selected based on Gumbel-Sigmoid is indeed effective.

Second, our method is more straightforward while achieving better results than the transformer-based two-stage model Deliberation networks~\citep{DBLP:conf/acl/IveMS19}. This result shows that our method is effective. However, the model structure is also as simple as possible compared to other methods, conducive to reproducibility, and easy to apply to other tasks.

Overall, our model maintains the best or near-best performance on the three test sets. Therefore, we reconfirmed the effectiveness and universality of Gumbel-Attention.

\subsection{Ablation Study}

\begin{table}
\centering
\setlength{\tabcolsep}{0.75mm}{
\small

\begin{tabular}{lcc}
\hline
  &BLEU  & METEOR \\
\hline
Gumbel-Attention Model       & 39.2  & 57.8     \\
Text-only Transformer       & 37.8  & 55.3     \\
\midrule 
-replace with vanilla-attention       & 38.6  & 55.9      \\
-replace with random image   & 38.3  & 55.6     \\
-shared parameters  & 38.5  & 56.0     \\
-w/o multi-modal gated fusion   & 38.9  & 56.2     \\
-w/o multi-modal similarity loss   & 39.0  & 57.1     \\
\hline
\end{tabular}}
\caption{Ablation study of our model on Test2016}
\label{tab:plain}
\end{table}

To investigate the effectiveness of different modules in Gumbel-Attention MMT, we further compare our method with the variants in Table 2.

\textbf{replace with vanilla-attention.} 
In this variant, we replace Gumbel-Attention with vanilla-attention, which performs a weighted sum of similarity between text and image information instead of selecting. This method can also make reasonable use of picture information, so the performance is improved compared with the text-only baseline. Additionally, the Gumbel-Attention model and vanilla-attention-based model have a gap of more than one bleu on Test2016, which demonstrates the influence of image noise on the MMT task.

\textbf{replace with random image.}
With reference to Multi-modal Transformer, we conducted random image replacement experiments.  
However, different from the previous conclusion, the result of row 4 shows that using random image to replace the original image in the Gumbel-Attention model is still better than text-only. 
This suggests that the random image as a regularization item improves the model effect, similar to the effect of random noise on the model~\citep{DBLP:journals/neco/Bishop95}.

\textbf{shared parameters.}
In this variant, we encode text representation and image-aware text representation with the shared encoder, which can reduce many parameters in the model. Although there is a decline compared with the original model, it is still an improvement over the text-only Transformer. We can obtain the gain brought by the image information with a nominal training cost through the method of sharing parameters.

\textbf{w/o multi-modal gated fusion.}
Instead of multi-modal gated fusion, we directly sum the outputs of two independent encoders. The result in row 6 shows that the multi-modal features fusion method based on the gate is a more suitable method than direct addition.

\textbf{w/o multi-modal similarity loss}
We delete the multi-modal similarity loss in the loss function. The results of row 7 show that the performance does not drop much compared with the original model. This result shows that image-aware representation is semantically close to text representation even without additional loss function constraints.

\section{Conclusion}
This paper has proposed Gumbel-Attention for multi-modal machine translation, aiming to reduce irrelevant noise in pictures by differentiable methods to select the image information. Our experiment results demonstrate the effectiveness of our model. We also introduced multi-modal similarity loss to restrict other image representation and text representation to be more similar. 

In future work, we would like to apply Gumbel Attention to other multi-modal tasks and investigate the theoretical interpretability of our method.

\bibliography{anthology,custom}
\bibliographystyle{acl_natbib}

\end{document}